\documentclass[letterpaper, 10 pt, conference]{ieeeconf}  

\IEEEoverridecommandlockouts                              

\usepackage{graphics} 
\usepackage{epsfig} 
\usepackage{mathptmx} 
\usepackage{times} 
\usepackage{amsmath} 
\usepackage{amssymb}  
\usepackage{cite}
\usepackage{multirow}

\title{\LARGE \bf
TASeg: Text-aware RGB-T Semantic Segmentation based on Fine-tuning Vision Foundation Models}

\author{Meng Yu$^{1}$, Te Cui$^{1}$, Qitong Chu$^{1}$, Wenjie Song$^{1}$, Yi Yang$^{1}$, Yufeng Yue$^{1}$
\thanks{This work is supported by the National Natural Science Foundation of China under Grant 62473050, 92370203.}
\thanks{$^{1}$Meng Yu, Te Cui, Qitong Chu, Wenjie Song, Yi Yang, and Yufeng Yue are with School of Automation, Beijing Institute of Technology, Beijing 100081, China. (Corresponding Author: Yufeng Yue, yueyufeng@bit.edu.cn)}%
}

\begin{document}

\maketitle
\thispagestyle{empty}
\pagestyle{empty}

\begin{abstract}

Reliable semantic segmentation of open environments is essential for intelligent systems, yet significant problems remain: 1) Existing RGB-T semantic segmentation models mainly rely on low-level visual features and lack high-level textual information, which struggle with accurate segmentation when categories share similar visual characteristics. 2) While SAM excels in instance-level segmentation, integrating it with thermal images and text is hindered by modality heterogeneity and computational inefficiency. To address these, we propose TASeg, a text-aware RGB-T segmentation framework by using Low-Rank Adaptation (LoRA) fine-tuning technology to adapt vision foundation models. Specifically, we propose a Dynamic Feature Fusion Module (DFFM) in the image encoder, which effectively merges features from multiple visual modalities while freezing SAM's original transformer blocks. Additionally, we incorporate CLIP-generated text embeddings in the mask decoder to enable semantic alignment, which further rectifies the classification error and improves the semantic understanding accuracy. Experimental results across diverse datasets demonstrate that our method achieves superior performance in challenging scenarios with fewer trainable parameters.

\end{abstract}

\section{INTRODUCTION}

Semantic segmentation plays a fundamental role in enabling scene understanding for intelligent systems. The accurate segmentation of images into semantically labeled regions allows these systems to interact with their environments more effectively, enhancing their operational capabilities and decision-making processes. 

Although prior works 
have achieved remarkable segmentation performance on standard RGB-based datasets, they often struggle in challenging conditions such as poor visibility caused by adverse weather or low illumination. To address this, researchers \cite{mfnet, pst900, LASnet, eaefnet, cainet} have introduced thermal/infrared images to enhance the performance of visual perception tasks. Despite significant progress, these methods predominantly rely on visual features, lacking the ability to comprehend semantic information. This limitation hampers their ability to accurately segment individual objects in complex scenes, particularly under occlusions, overlapping objects, or ambiguous boundaries. These limitations highlight the need for incorporating additional semantic cues, such as textual information, to help improve high-level semantic understanding for fine-grained segmentation.

\begin{figure}[t!]
\centerline{\includegraphics[width=0.8\columnwidth]{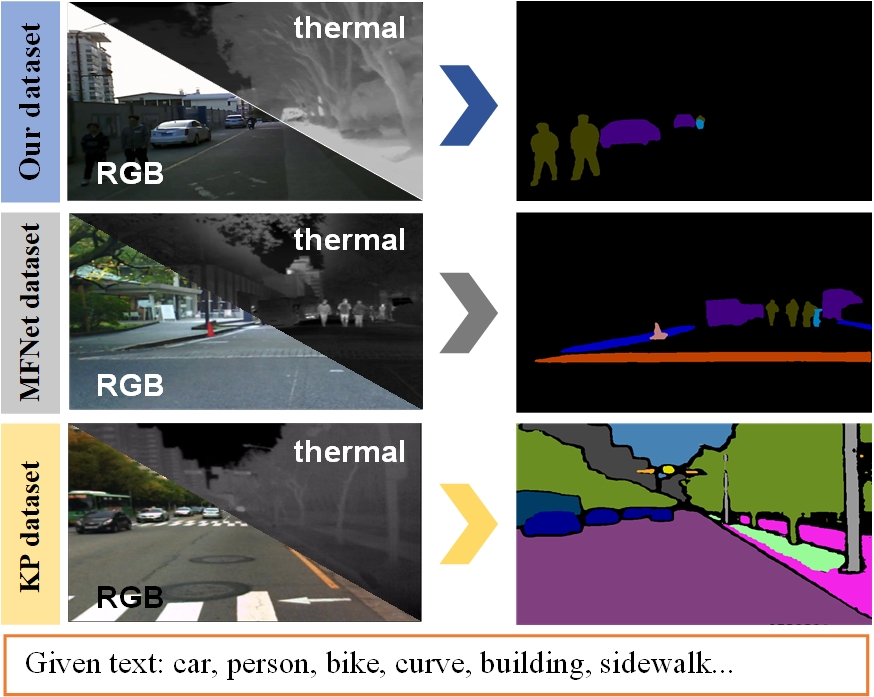}}
\caption{We introduce \textbf{TASeg}, a text-aware segmentation framework, which facilitates RGB-T semantic segmentation across various scenarios.}
\vspace{-1.5em}
\label{fig1}
\end{figure}

Recent advancements in foundation models, such as Segment Anything Model (SAM) \cite{sam}, have demonstrated exceptional generalization capabilities and strong instance segmentation capabilities. Similarly, Vision-Language Models (VLMs) such as CLIP \cite{clip} and ALIGN \cite{align} have revolutionized semantic understanding by aligning visual and textual representations \cite{ovseg, groundsam}. However, a significant issue arises when applying these models to RGB-T domain: the inherent heterogeneity between RGB and thermal modalities often leads to suboptimal performance. This occurs because the data distributions in the RGB-T domain deviate from those in the models' pretraining data. Additionally, fine-tuning these large models with full parameters is computationally expensive and often impractical, especially when task-specific datasets are limited in size. 

Motivated by the above observations, we propose a novel framework that adapts SAM for RGB-T semantic segmentation tasks. Specifically, Low-Rank Adaptation (LoRA) fine-tuning technology is employed to efficiently adapt the image encoder and mask decoder of SAM, which introduces trainable low-rank matrices into the model, enabling task-specific adaptation while keeping the majority of the pretrained parameters frozen. Additionally, we introduce a Dynamic Feature Fusion Module (DFFM) in the image encoder, which effectively merges features from multiple visual modalities. Furthermore, textual guidance using pretrained CLIP model is incorporated within the mask decoder to enhance semantic comprehension. By aligning text embeddings with visual features, our method rectifies semantic classification errors and improves the object segmentation accuracy.

\begin{figure*}[ht!]
\centerline{\includegraphics[width=\textwidth]{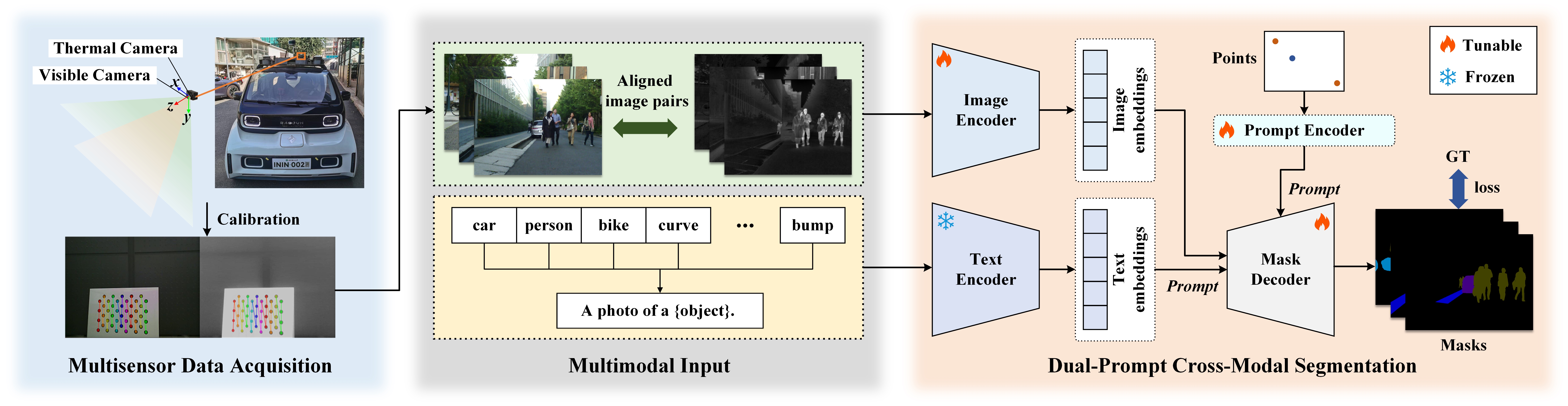}}
\caption{The overall framework of our TASeg, which processes pixel-aligned RGB-thermal image pairs along with predefined semantic classes. }
\vspace{-1.5em}
\label{fig2}
\end{figure*}

In summary, the contributions of this paper are summarized as follows:
\begin{enumerate}
    \item We propose TASeg, which employs LoRA to adapt SAM for RGB-T semantic segmentation, where a Dynamic Feature Fusion Module is presented to enable effective RGB-T feature fusion.
    \item We introduce text-guided mask generation using CLIP, enabling semantic alignment and improving segmentation accuracy.
    \item Extensive experiments on benchmark datasets demonstrate the effectiveness of our approach, showcasing its ability to generalize across diverse data distributions.
\end{enumerate}

\section{RELATED WORKS}
\subsection{RGB-T Semantic Segmentation}
Single-modal segmentation networks 
struggle in complex scenarios, driving the development of RGB-T fusion methods. Early works like MFNet \cite{mfnet} and RTFNet \cite{rtfnet} employed dual-stream architectures with element-wise summation for feature fusion.
However, these simplistic fusion strategies overlooked cross-modal discrepancies. Subsequent studies introduced refined mechanisms: GMNet \cite{gmnet} optimized features through channel/spatial attention, and EGFNet \cite{egfnet} preserved boundaries using edge priors.
Recent approaches like EAEFNet \cite{eaefnet} further explored explicit complementary modeling and morphological feature extraction. Despite progress, they often face challenges in accurately understanding and segmenting individual objects within complex images containing occlusions, overlapping objects, or ambiguous boundaries. Consequently, our work is motivated by the need to develop a framework that effectively preserves the integrity of individual objects.

\subsection{VLMs-based Semantic Segmentation}
Foundation models like SAM and CLIP have revolutionized open-vocabulary segmentation. 
OpenSeg \cite{openseg} improved granularity via region-word grounding. 
Two-stage methods generated class-agnostic masks followed by CLIP-based classification, whereas Ovseg \cite{ovseg} fine-tuned CLIP for enhanced performance. Grounded-SAM \cite{groundsam} combined open-set detection with segmentation for arbitrary text inputs. However, they primarily focus on clean RGB imagery and may struggle with domain-specific challenges. In particular, the inherent heterogeneity between modalities and the lack of thermal-specific pretraining limit their effectiveness in specialized domains such as thermal imaging. 

This gap motivates our exploration of domain-adaptive segmentation approaches that can bridge the modality gap while preserving the generalization capabilities of foundation models. Parameter-efficient fine-tuning (PEFT) techniques have been developed for adapting large-scale models to specific downstream tasks. For example, SAM-Adapter \cite{sam-adapter} inserted lightweight adapter \cite{adapter} layers, while SAMed \cite{samed} adopted LoRA to update weights via low-rank decomposition \cite{lora}. Among these methods, LoRA stands out for its balance between parameter efficiency and performance, making it an ideal choice for our work.

\section{METHOD}
\subsection{Framework Overview}
The proposed framework, depicted in Fig. \ref{fig2}, processes pixel-aligned RGB-thermal image pairs along with predefined semantic classes. First, both modalities are jointly encoded: the RGB-thermal images are fed into a tunable image encoder to extract hierarchical visual features, while the semantic class labels are transformed into text embeddings via a text encoder. Subsequently, these cross-modal representations are fused in a dual-prompt segmentation stage, where the mask decoder dynamically integrates text prompts and spatial prompts (points) to guide the segmentation process. 

\subsection{Dual-Prompt Cross-Modal Segmentation}

\subsubsection{\textbf{Fine-tuning the Image Encoder}}
\begin{figure}[t!]
\centerline{\includegraphics[width=0.9\columnwidth]{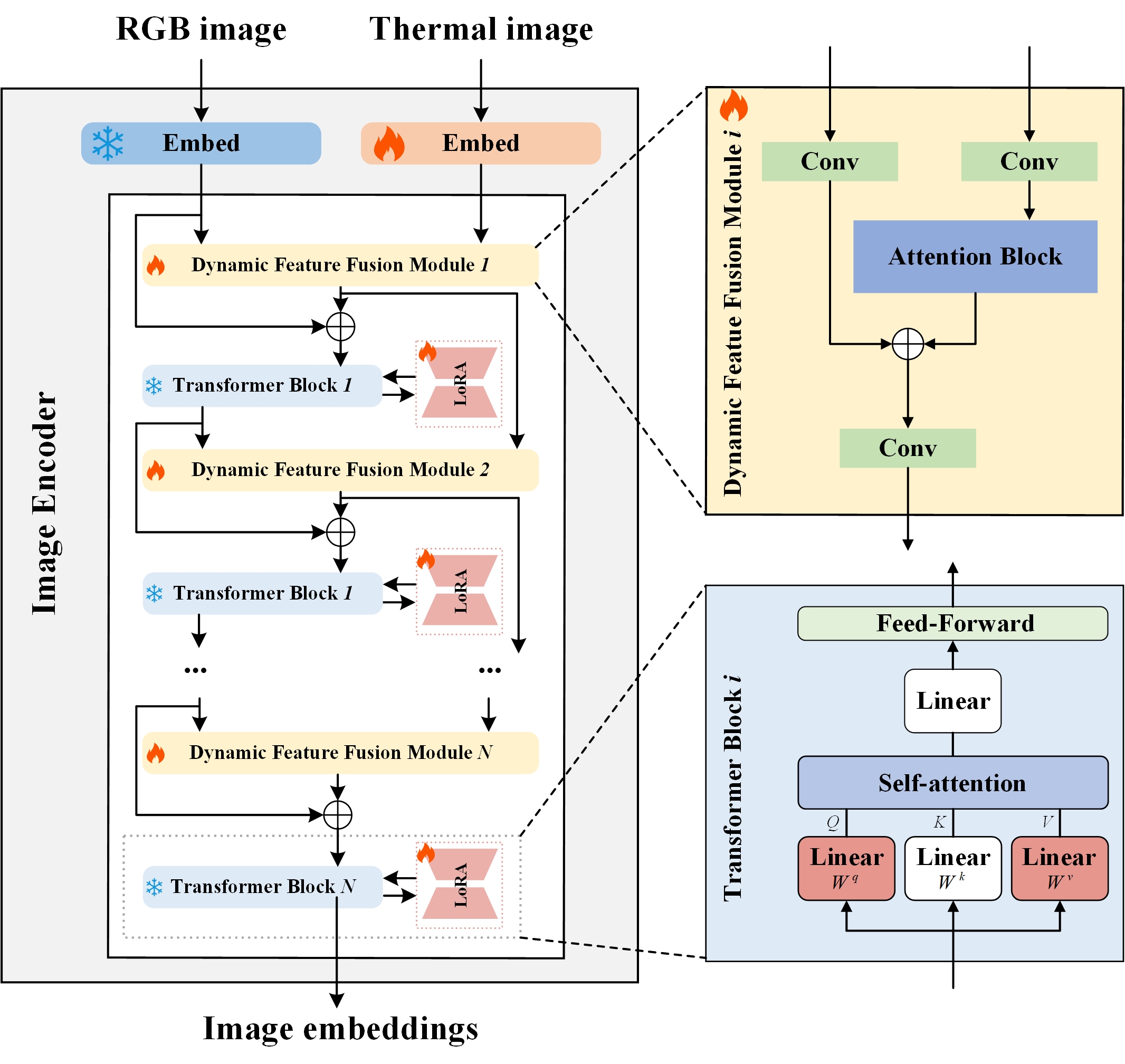}}
\caption{The scheme of fine-tuning the Image Encoder.}
\vspace{-1.5em}
\label{fig3}
\end{figure}

While SAM is originally designed for RGB modality, our task focuses on RGB-T semantic segmentation, necessitating a mechanism to effectively fuse multisensor features. To address this, we propose the Dynamic Feature Fusion Module (DFFM), which builds upon the frozen transformer blocks of SAM while introducing trainable components to enable RGB-T feature fusion. The structure is presented in Fig. \ref{fig3}.

Specifically, the input RGB and thermal images are denoted as $\textbf{I}_{rgb}$ and $\textbf{I}_{th}$, respectively. These images are firstly converted into patch embedding sequences $\textbf{e}_{rgb}$ and $\textbf{e}_{th}$ using the PatchEmbed module, which is represented as:
\begin{equation}
\textbf{e}_{rgb}=PatchEmbed(\textbf{I}_{rgb}),\textbf{e}_{th}=PatchEmbed^*(\textbf{I}_{th}),
\label{eq1}
\end{equation}

The PatchEmbed module splits each image into non-overlapping patches via convolutional operations and projects these patches into an embedded space. Notably, while the RGB branch remains frozen and the thermal branch is trainable.

Then, the embedded features are transferred into the Dynamic Feature Fusion Module. For the $i$-th DFFM and Transformer Block, let $\textbf{F}_{DFFM}^{i-1}$ and $\textbf{F}_{TB}^{i-1}$ denote the outputs of the ($i$-1)-th DFFM and TB, respectively. The $i$-th output of DFFM is computed as:
\begin{equation}
\textbf{F}_{DFFM}^{i}=conv(conv(\textbf{F}_{DFFM}^{i-1})+AttentionBlock(conv(\textbf{F}_{TB}^{i-1})))
\label{eq2}
\end{equation}
where \textit{conv} represents a $1\times1$ convolution operation used to project the input features to a consistent dimensionality. The Attention Block adopts the Squeeze-and-Excitation Block (SEBlock) \cite{seblock}, which is particularly effective as it adaptively recalibrates channel-wise feature responses by explicitly modeling interdependencies between channels.  

The output of the $i$-th TB, denoted as $\textbf{F}_{TB}^{i}$, is computed as:
\begin{equation}
\textbf{F}_{TB}^{i}=TransformerBlock(\textbf{F}_{TB}^{i-1}+\textbf{F}_{DFFM}^{i})
\label{eq3}
\end{equation}
where TransformerBlock refers to the frozen transformer block from SAM. By adding the outputs of the ($i$-1)-th TB and the $i$-th DFFM, we enhance the feature representation while preserving the pretrained knowledge encoded in the frozen transformer blocks.

For the first DFFM ($i$=1), the input features are initialized as the patch embedding sequences of the RGB and thermal images:
\begin{equation}
\textbf{F}_{DFFM}^{0}=\textbf{e}_{th}, \textbf{F}_{TB}^{0}=\textbf{e}_{rgb}
\label{eq4}
\end{equation}
where this initialization ensures that the multisensor features are effectively integrated from the beginning of the network.

By iteratively refining the feature representations, the Image Encoder strikes a balance between preserving the generalization capabilities of the pretrained SAM and adapting to the specific requirements of RGB-T data.

To adapt the pretrained SAM model to RGB-T semantic segmentation task, LoRA fine-tuning is employed in the image encoder. LoRA fine-tunes the model by introducing low-rank updates to the pretrained parameters, its scheme applied in the transformer block is shown in Fig. \ref{fig3}. The core of the attention mechanism in the Transformer Blocks involves computing a weighted combination of values $V$ based on the similarity between queries $Q$ and keys $K$. When fine-tuning, we need to focus on the parts that most significantly impact the model's output and performance. Specifically, the $Q$ and $V$ layers are critical because they directly influence the computation of attention weights and the final output representation. By adjusting these two parameters, the maximum performance improvement can be obtained. Mathematically, this process can be expressed as:
\begin{equation}
\begin{aligned}
& Q=W^q\textbf{F}=W_0^q\textbf{F}+B^qA^q\textbf{F} \\
& V=W^v\textbf{F}=W_0^v\textbf{F}+B^vA^v\textbf{F}
\label{eq5}
\end{aligned}
\end{equation}
where $W_0^q, W_0^v\in \mathbb{R}^{d\times d}$ are the pretrained frozen weights of SAM, $B^q, B^v \in \mathbb{R}^{d\times r}$ and $A^q,A^v \in \mathbb{R}^{r\times d}$ are trainable low-rank matrices, with $r$ being the rank of the decomposition. $\textbf{F}$ represents the input feature to the Transformer Blocks.

\subsubsection{\textbf{Text-aware Mask Generation}}
The original mask decoder in SAM takes several inputs to generate segmentation masks: image embeddings $\textbf{e}_{en}$, image positional encoding $\textbf{e}_{pe}$, sparse embeddings $\textbf{e}_{s}$, and dense embeddings $\textbf{e}_{d}$. Here, $\textbf{e}_{pe}$ provides position information, which is crucial for aligning features in the image space. Then, the transformer-based mask decoder processes three key components:
\begin{equation}
\textbf{E}_s=\textbf{e}_{en}+\textbf{e}_{d},\textbf{E}_p=\textbf{e}_{pe},\textbf{E}_t=concat(\textbf{e}_{iou},\textbf{e}_{mask},\textbf{e}_{s})
\label{eq6}
\end{equation}
where $\textbf{e}_{iou}$ and $\textbf{e}_{mask}$ represent iou tokens and mask tokens, respectively. The transformer block processes these components to produce two outputs:
\begin{equation}
\textbf{e}_m,\textbf{e}_f=Transformer(\textbf{E}_s,\textbf{E}_p,\textbf{E}_t)
\label{eq7}
\end{equation}

To adapt the feature in the mask decoder, LoRA is also applied to the transformer layers and token-to-image attention. Through upscaling operations, the final image mask embeddings $\textbf{e}_{M}$ can be obtained from $\textbf{e}_{m}$. 

As SAM lacks explicit semantic understanding, which may lead to errors in semantic classification. In our work, we introduce textual information using the CLIP model. Specifically, we utilize CLIP to generate text embeddings $\textbf{e}_{t}$ corresponding to the semantic classes. 

Then, we use image mask embeddings $\textbf{e}_{M}$ as query vectors, while the text embeddings $\textbf{e}_{t}$ provide key-value pairs through linear transformations. This allows us to compute the aligned text-image feature as follows:
\begin{equation}
\textbf{F}_M=\text{Softmax}\left(\frac{\textbf{e}_{M}\textbf{W}_Q(\textbf{e}_{t}\textbf{W}_K)^T}{\sqrt{d_k}}\right)\textbf{e}_{t}^T\textbf{W}_V
\label{eq8}
\end{equation}
where $\textbf{W}_Q$, $\textbf{W}_K$, $\textbf{W}_V$ are learnable weights matrices. The scaling factor $\sqrt{d_k}$ ensures gradient stability during training. Finally, the predicted masks can be obtained from the resulting features through a projection.

\subsubsection{\textbf{Supervised Loss Function}}
To optimize our model effectively, we utilize Cross-Entropy Loss ($\mathcal{L}_{1}$) to ensure pixel-wise accuracy. To address the challenges of fuzzy boundaries and class imbalance, we introduce the Dice Loss ($\mathcal{L}_{Dice}$), which is particularly effective for tasks where boundary regions are ambiguous.

\begin{figure*}[ht!]
\centerline{\includegraphics[width=\textwidth]{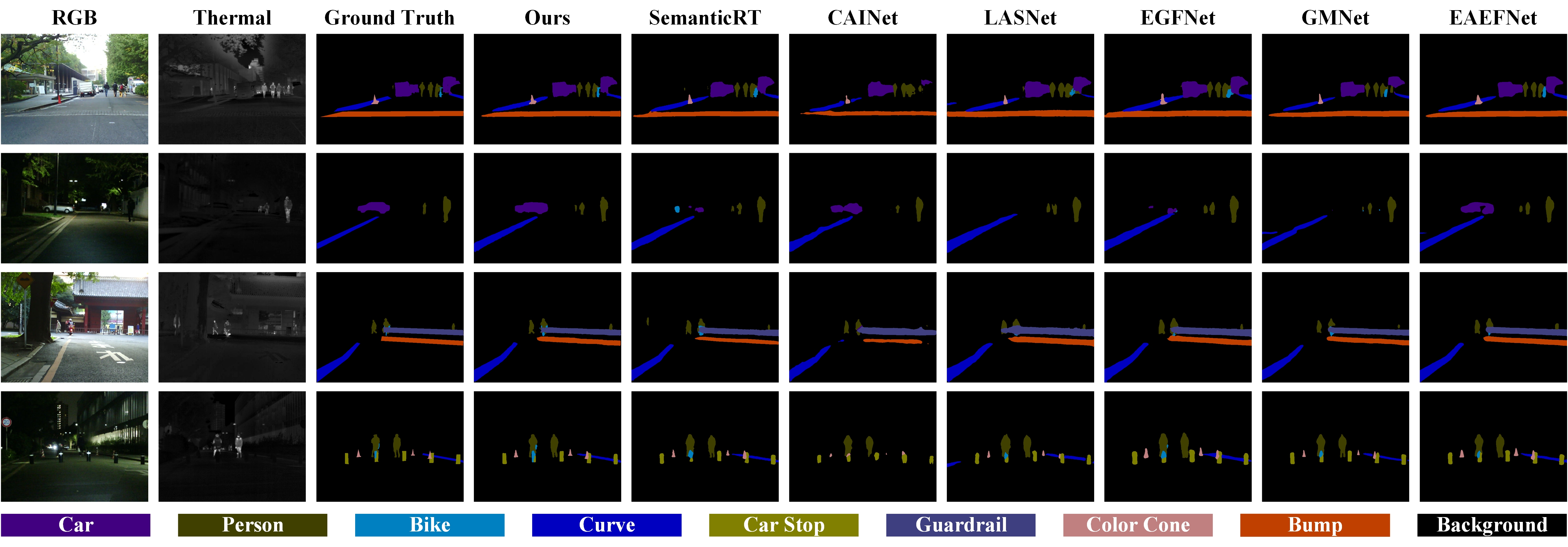}}
\vspace{-0.5em}
\caption{The sample qualitative results on MFNet dataset.}
\vspace{-1.5em}
\label{fig4}
\end{figure*}

\begin{table}
\caption{Quantitative comparisons (\%) on MFNet and PST900 evaluation set. The best result is highlighted in \textbf{Bold}.}
\setlength{\tabcolsep}{3.2pt}
\centering
\begin{tabular}{cccccc} 
\hline
\multirow{2}{*}{Methods} & {Trainable } & \multicolumn{3}{c}{MFNet}     & \multirow{2}{*}{PST900}  \\
& params/M & Daytime & Nighttime & Overall &  \\ 
\hline
MFNet \cite{mfnet}  & 8.4   & 36.1    & 36.8      & 39.7    & 57.02   \\
RTFNet \cite{rtfnet} & 245.7  & 45.8    & 54.8  & 53.2    & 60.46   \\
PSTNet \cite{pst900} & 105.8  & -       &   & 48.4    & 68.36    \\
GMNet \cite{gmnet}  & 149.8   & 49.0    & 57.7    & 57.3    & 84.12  \\
EGFNet \cite{egfnet} & 201.3  & 47.3    & 55.0    & 54.8    & 78.51 \\
FDCNet \cite{fdcnet} & 52.91    & 47.8    & 56.8   & 56.3    & 77.11    \\
EAEFNet \cite{eaefnet} & 200.4    & -   &   & 58.9    & 85.40     \\
semanticRT \cite{semanticrt}  & -  & -   &    & 58.0    & 84.47    \\
LASNet \cite{LASnet}  & 93.58  & 45.2    & 58.7   & 54.9    & 84.40 \\
MDRNET+ \cite{mdrnet} & 64.6 & 48.1    & 56.7    & 56.8    & 74.62    \\
CAINet \cite{cainet} & 12.16  & -  &   & 58.6    & 84.74   \\
MDNet \cite{mdnet} & 26.96  & 59.8    & 48.7  & 58.9    & 82.97   \\ 
\hline
Ours & 28.77  & \textbf{82.4}  & \textbf{71.8}  & \textbf{77.6}  & \textbf{86.09}  \\
\hline
\end{tabular}
\vspace{-1.5em}
\label{tab1}
\end{table}

\section{EXPERIMENTS}

\subsection{Experimental Setup}
We conduct comprehensive evaluations on four publicly RGB-T benchmarks: MFNet \cite{mfnet}, PST900 \cite{pst900}, KAIST Pedestrain (KP) \cite{uda}, FMB \cite{FMB}, and our self-collected dataset in real-world scenes. 

We adopt mIoU (mean Intersection over Union) as the primary metric, complemented by per-class IoU analysis.

Our framework builds on SAM's ViT-H image encoder and mask decoder, both frozen and adapted via LoRA for parameter-efficient tuning. Additionally, the text encoder is initialized with the pretrained CLIP model based on the ViT-L/16 backbone. We train our model using an NVIDIA GTX 3090 GPU, with an initial learning rate of $5 \times 10^{-4}$ and a weight decay of 0.01. During the testing phase, the model trained on the KP dataset is directly applied to evaluate the FMB dataset without further training.

\subsection{Comparison with State-of-the-art Methods}
\subsubsection{\textbf{Evaluation on MFNet and PST900 datasets}}
We compare our method with those of leading SOTA methods.


\begin{figure*}[ht!]
\centerline{\includegraphics[width=0.8\textwidth]{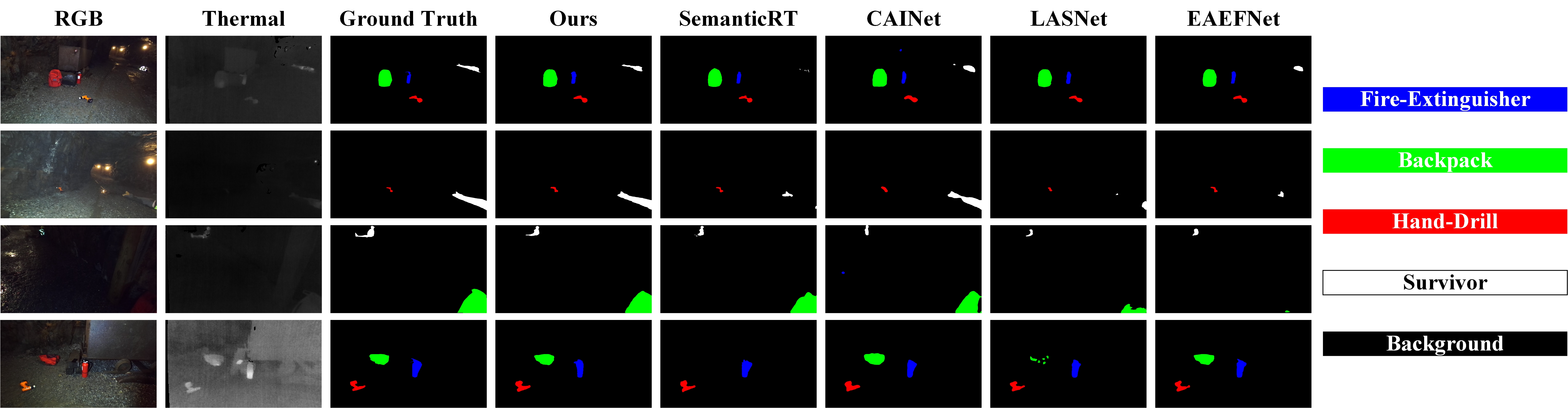}}
\vspace{-0.5em}
\caption{The sample qualitative results on PST900 dataset.}
\label{fig5}
\end{figure*}

\begin{table*}
\caption{Quantitative comparisons (\%) on KP dataset evaluation set. The best result under each case is highlighted in \textbf{Bold}.}
\setlength{\tabcolsep}{4pt}
\renewcommand{\arraystretch}{1}
\centering
\begin{tabular}{ccccccccccccccccccccc} 
\hline
Methods & \rotatebox{90}{Road} & \rotatebox{90}{Sidewalk} & \rotatebox{90}{Building} & \rotatebox{90}{Wall} & \rotatebox{90}{Fence} & \rotatebox{90}{Pole} & \rotatebox{90}{Traffic light} & \rotatebox{90}{Traffic sign} & \rotatebox{90}{Vegetation} & \rotatebox{90}{Terrain} & \rotatebox{90}{Sky} & \rotatebox{90}{Person} & \rotatebox{90}{Rider} & \rotatebox{90}{Car} & \rotatebox{90}{Truck} & \rotatebox{90}{Bus} & \rotatebox{90}{Train} & \rotatebox{90}{Motorcycle} & \rotatebox{90}{Bicycle} & mIoU \\
\hline
MFNet \cite{mfnet}  & 93.5 & 23.6  & 75.1  & 0.0  & 0.1   & 9.1  & 0.0 & 0.0  & 69.3 & 0.2 & 90.4 & 24.0 & 0.0 & 69.6 & 0.3  & 0.3 & 0.0 & 0.0 & 0.6 & 24.0    \\
RTFNet \cite{rtfnet} & 94.6 & 39.4 & 86.6 & 0.0 & 0.6  & 0.0 & 0.0  & 0.0  & 81.7 & 3.7 & 92.8 & 58.4 & 0.0 & 87.7 & 0.0 & 0.0 & 0.0  & 0.0  & 0.5 & 28.7  \\
UDA \cite{uda} & 95.7 & 30.4 & 76.0 & \textbf{0.2} & 18.9 & 10.6 & 0.0  & 11.4 & 76.2 & 18.8 & \textbf{{93.8}} & 62.9 & 0.1 & 69.6 & 2.0 & 42.0 & 0.0  & 4.9 & 20.2 & 33.4  \\
CMX \cite{cmx} & {97.7} & \textbf{{53.8}} & \textbf{{90.2}} & 0.0  & 47.1  & \textbf{{46.2}} & 10.9 & \textbf{{45.1}} & \textbf{{87.2}} & \textbf{{34.3}} & 93.5 & 74.5 & 0.0  & 91.6 & 0.0  & 59.7 & 0.0 & 46.1 & 0.2 & 46.2  \\ 
CAINet \cite{cainet} & 95.3 & 44.7 & 79.4 & 0.0 & 44.7 & 20.0 & 8.1 & 28.7 & 63.1 & 16.0 & 88.5 & 56.3 & 0.0 & 83.4 & 4.6 & 58.8 & 0.0 & 28.1 & 0.3 & 40.6 \\
\hline
Ours & \textbf{{97.8}} & 47.5  & 83.4 & 0.0  & \textbf{{54.3}}  & 42.5 & \textbf{26.8} & 39.7 & 64.0 & 26.5 & 92.2 & \textbf{{79.0}} & \textbf{{26.2}}  & \textbf{{92.7}} & \textbf{{11.5}}  & \textbf{{77.9}} & 0.0 & \textbf{{48.2}} & \textbf{28.3} & \textbf{{51.1}}  \\
\hline
\end{tabular}
\vspace{-1.5em}
\label{tab2}
\end{table*}

The quantitative results are shown in Table \ref{tab1}, it is noteworthy that our method exhibited superior performance compared to its competitors, achieving an overall mIoU of 77.6\% and 86.09\% on the MFNet and PST900 dataset, respectively. Besides, our method achieved the highest performance in both daytime and nighttime scenarios. 

Qualitative semantic segmentation results on the MFNet dataset are illustrated in Fig. \ref{fig4}. In contrast, our approach effectively segments objects with precision and distinct boundaries, even at a distance. This precision is attributed to the proposed Dynamic Feature Fusion Module, which effectively preserves critical image features during multimodal fusion. 


Similarly, the segmentation results on the PST900 dataset are presented in Fig. \ref{fig5}. For example, in the first three rows, other methods either fail to completely segment or misdetect the ``Survivor" class, whereas our method maintains robust segmentation performance.

\subsubsection{\textbf{Evaluation on KP and FMB datasets}}
We compare our method with five publicly available methods, including MFNet \cite{mfnet}, RTFNet \cite{rtfnet}, UDA \cite{uda}, CMX \cite{cmx}, and CAINet \cite{cainet}. 

The quantitative results are summarized in Table \ref{tab2}. Notably, our method achieves an overall performance improvement of 4.9\% over the second-best method. Qualitative visualizations of four representative samples are presented in Fig. \ref{fig6}. The results clearly show that our method's segmentation performance aligns more closely with the ground truth annotations. For instance, our method accurately segment the ``Traffic light" and ``Traffic sign" classes in all cases. Additionally, in the second row, only our method successfully segment the ``Terrain" class. Overall, our method achieves superior segmentation performance, demonstrating its ability to handle complex and diverse scenarios effectively.

To evaluate the generalization capability of our model, we directly apply the model trained on the KP dataset to the FMB test dataset. The last two rows of Fig. \ref{fig6} illustrate the visualization results under nighttime and foggy conditions. Compared to the other methods, our method effectively handles these challenging scenarios, accurately segmenting each category with precise boundaries. This highlights the strong robustness and generalization ability of our approach, even under adverse conditions.

\begin{figure}[ht!]
\centerline{\includegraphics[width=\columnwidth]{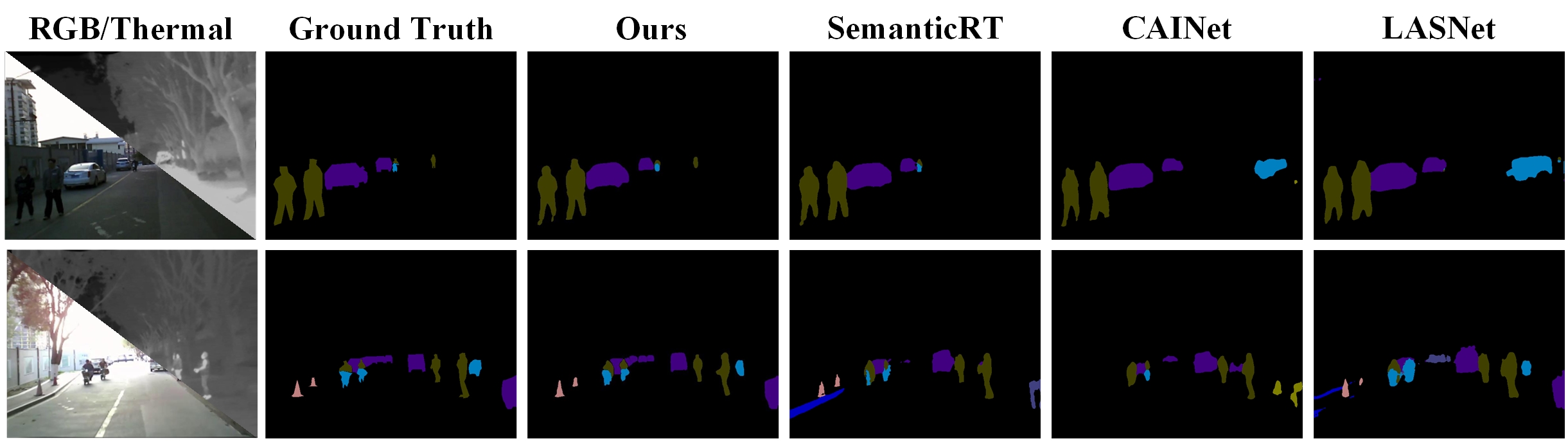}}
\caption{The sample qualitative results on our dataset.}
\vspace{-1.5em}
\label{fig7}
\end{figure}

\subsubsection{\textbf{Evaluation on our dataset}}
To further validate the practical applicability of our method, we evaluate several approaches using our dataset in real-world scenarios. All methods are tested using their pre-trained models from the MFNet dataset. Qualitative results are presented in Fig. \ref{fig7}. In contrast, our method consistently achieves precise and robust segmentation, even in complex and dynamic environments. 


\subsection{Complexity Analysis}
In addition to leveraging frozen parameters from the large model, our method significantly reduces the total number of trainable parameters to 28.77M, seen in Table \ref{tab1}. Furthermore, our approach is computationally efficient, as it can be trained on a single 3090 GPU without demanding extensive computational resources, unlike many VLMs-based methods that rely on high computational power. During inference, our method achieves a speed of 4.65 fps for images with a resolution of 640 $\times$ 480. While the real-time performance is not yet optimal, the trade-off is justified by the superior segmentation accuracy and robustness, which outperform those of competing methods.

\subsection{Ablation Studies}
We perform ablation studies on the MFNet dataset, and the quantitative results are presented in Table \ref{tab3}. For the ``Baseline" condition, we retain LoRA layers in the image encoder of SAM, where RGB and thermal image embeddings are simply concatenated. The results demonstrate that each module significantly enhances the performance of segmentation. For instance, the introduction of the DFFM improves the mIoU from 71.6 \% (baseline) to 73.8 \%, highlighting its ability to effectively fuse multimodal features and refine feature representations. Adding LoRA layers to the mask decoder further boosts performance, increasing the mIoU to 72.7 \%, and to 75.1 \% when integrated with DFFM. Furthermore, the incorporation of text information further improves semantic understanding, yielding an improvement of 1.5 \% over the configuration with only the DFFM.

\begin{table}
\caption{Ablation studies on MFNet dataset. The best one is in \textbf{Bold}.}
\renewcommand{\arraystretch}{1}
\centering
\begin{tabular}{cccccc} 
\hline
No. & Baseline & DFFM & LoRA in mask decoder & Text & mIoU  \\ 
\hline
1 & \checkmark  &     &     &      & 71.6  \\
2 & \checkmark  &     & \checkmark    &      & 72.7  \\
3 & \checkmark  &     & \checkmark    & \checkmark    & 74.9  \\
4 & \checkmark  & \checkmark   &  &   & 73.8  \\
5 & \checkmark  & \checkmark   &  & \checkmark    & 75.3  \\
6 & \checkmark  & \checkmark   & \checkmark &      & 75.1  \\ 
\hline
7 & \checkmark & \checkmark   & \checkmark & \checkmark  & \textbf{77.6}  \\
\hline
\end{tabular}
\vspace{-1.5em}
\label{tab3}
\end{table}

\begin{figure*}[t!]
\centerline{\includegraphics[width=0.7\textwidth]{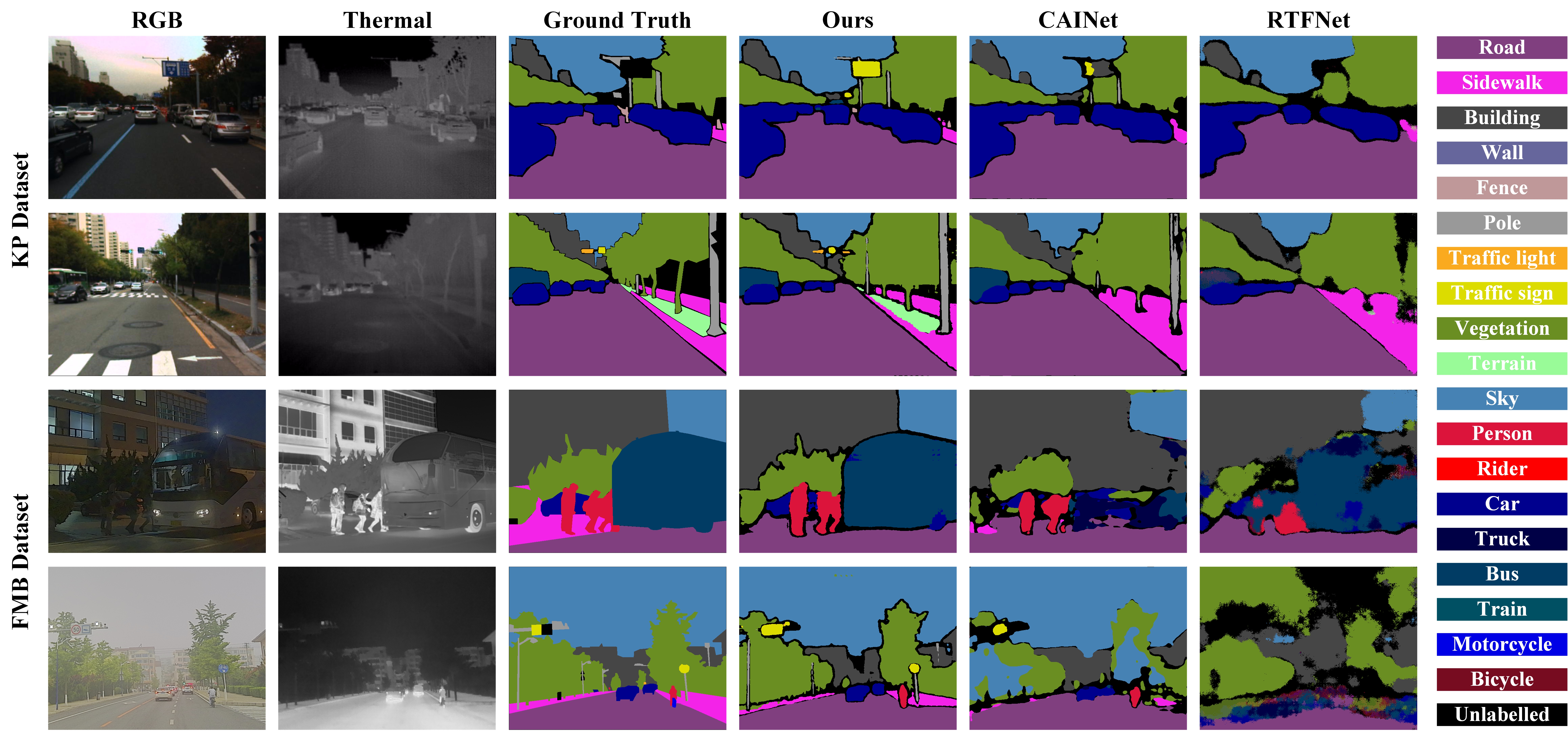}}
\caption{The sample qualitative results on KP and FMB dataset.}
\vspace{-1.5em}
\label{fig6}
\end{figure*}

\section{CONCLUSIONS}
In this paper, we propose TASeg, a novel framework that adapts SAM for RGB-T semantic segmentation. By freezing the majority of SAM's original parameters and selectively introducing trainable LoRA layers, our approach achieves a balance between computational efficiency and task-specific performance, making it well-suited for multimodal tasks. Compared to existing methods, our method is capable of segmenting pixel-level objects and efficiently adapts to diverse data distributions through efficient fine-tuning. Overall, our method addresses the issues of both segmentation integrity and modality heterogeneity. Extensive experiments validate the effectiveness and generalization of our work.



\bibliographystyle {unsrt} 
\bibliography{reference}

\end{document}